\documentclass[journal]{IEEEtran}
\usepackage{amsmath,amsfonts}
\usepackage[linesnumbered,ruled,vlined]{algorithm2e}
\usepackage{array}
\usepackage{textcomp}
\usepackage{stfloats}
\usepackage{url}
\usepackage{verbatim}
\usepackage{graphicx}
\usepackage{cite}
\usepackage{hyperref}
\usepackage{booktabs}
\usepackage{multirow}
\usepackage{makecell}
\usepackage{diagbox}
\usepackage[caption=false]{subfig}

\newcolumntype{L}{>{\ttfamily}l}
\newcommand{\multirowcenter}[3]{\multirow{#1}{#2}[-2pt]{#3}}
\newcounter{magicrownumbers}
\newcommand\rownumber{\stepcounter{magicrownumbers}\arabic{magicrownumbers}}

\begin{document}

\title{NavComposer: Composing Language Instructions for Navigation Trajectories through Action-Scene-Object Modularization}

\author{Zongtao He,
        Liuyi Wang,
        Lu Chen,
        Chengju Liu,
        and~Qijun~Chen,~\IEEEmembership{Senior~Member,~IEEE}
\thanks{This paper is supported by the National Natural Science Foundation of China under Grants 62473295 and 62233013. (Corresponding author: Chengju~Liu, Qijun~Chen)}
\thanks{The authors are with the Department of Control Science and Engineering, Tongji University, Shanghai, 201804, China. E-mail: xingchen327@tongji.edu.cn, wly@tongji.edu.cn, chenlu\_i@tongji.edu.cn, liuchengju@tongji.edu.cn, qjchen@tongji.edu.cn.}%
\thanks{We will release our code at https://github.com/RavenKiller/NavComposer.}
}

\markboth{Preprint version}%
{He \MakeLowercase{\textit{et al.}}: NavComposer: Composing Language Instructions for Navigation Trajectories through Action-Scene-Object Modularization}


\maketitle

\begin{abstract}
Language-guided navigation is a cornerstone of embodied AI, enabling agents to interpret language instructions and navigate complex environments.
However, expert-provided instructions are limited in quantity, while synthesized annotations often lack quality, making them insufficient for large-scale research.
To address this, we propose NavComposer, a novel framework for automatically generating high-quality navigation instructions.
NavComposer explicitly decomposes semantic entities such as actions, scenes, and objects, and recomposes them into natural language instructions.
Its modular architecture allows flexible integration of state-of-the-art techniques, while the explicit use of semantic entities enhances both the richness and accuracy of instructions.
Moreover, it operates in a data-agnostic manner, supporting adaptation to diverse navigation trajectories without domain-specific training.
Complementing NavComposer, we introduce NavInstrCritic, a comprehensive annotation-free evaluation system that assesses navigation instructions on three dimensions: contrastive matching, semantic consistency, and linguistic diversity.
NavInstrCritic provides a holistic evaluation of instruction quality, addressing limitations of traditional metrics that rely heavily on expert annotations.
By decoupling instruction generation and evaluation from specific navigation agents, our method enables more scalable and generalizable research. 
Extensive experiments provide direct and practical evidence for the effectiveness of our method.
\end{abstract}


\section{Introduction}

Embodied AI \cite{duan2022survey, li2023behavior, Liu2024AligningCS} has emerged as a critical frontier in robotics and artificial intelligence, with language playing a pivotal role in facilitating intuitive human-robot interaction.
Among its various applications, language-guided navigation, such as Vision-and-Language Navigation (VLN) \cite{anderson8578485}, focuses on developing agents capable of navigating complex environments using only natural language instructions and visual observations.
Training such agents using deep neural networks requires large-scale, high-quality annotated datasets.
Early VLN research \cite{anderson8578485, ku2020room, vasudevan2021talk2nav, qi9156641} relied on crowd-sourced tools to collect expert annotations for navigation instructions.
However, the complexity of navigation tasks makes obtaining sufficient annotations prohibitively expensive and time-consuming.

\begin{figure}[tbp]
    \centering
    \includegraphics[width=0.48\textwidth]{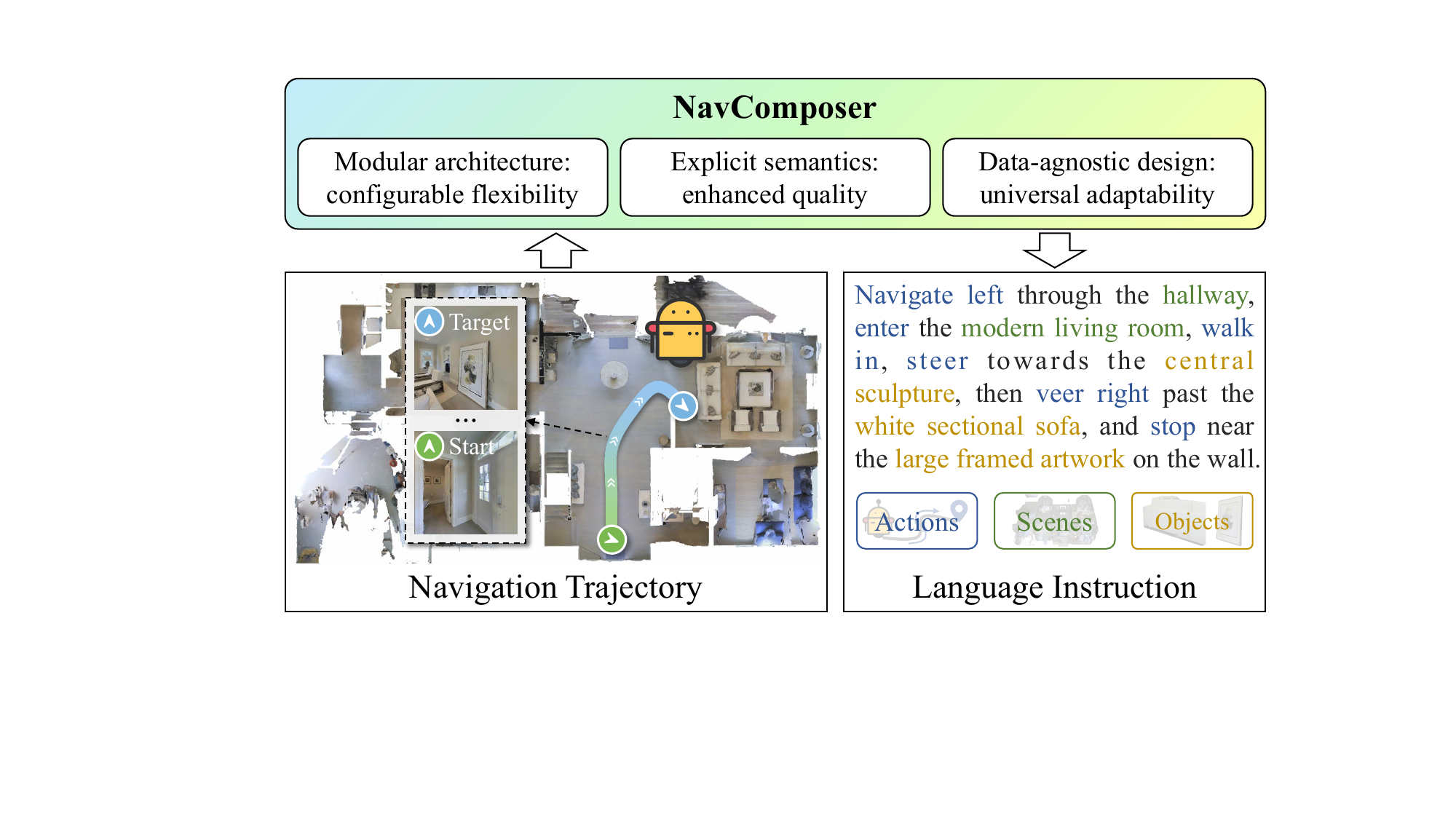}
    \caption{Overview of our method. NavComposer extracts semantic entities from navigation trajectories and synthesizes concise, natural language instructions.}
    \label{fig:introduction}
\end{figure}
The high cost of annotation has driven the development of data augmentation.
Although existing methods have made significant progress \cite{fried2018speaker, tan2019learning, agarwal2019visual, magassouba2021crossmap, wu2021improved, wang2022ressts}, they still face critical challenges that limit their effectiveness and general applicability.
Most current methods use end-to-end deep neural networks to directly map images to text \cite{fried2018speaker, agarwal2019visual}.
These models offer little control over the instruction generation process and lack the flexibility needed to integrate advanced techniques like large language models (LLMs).
Second, the consistency and diversity of generated instructions remain insufficient.
For instance, instructions from the LSTM-based model \cite{fried2018speaker} often omit key navigation landmarks and exhibit repetitive phrasing and structural biases.
Moreover, most existing methods are tightly coupled to specific datasets in both data structure (panoramic vision in R2R \cite{anderson8578485}) and data domain (reconstructed indoor villa scenarios).
As a result, these methods struggle to generalize to new environments or tasks.

We propose NavComposer, a novel modular framework for generating high-quality language instructions for any navigation trajectory, as illustrated in Fig. \ref{fig:introduction}.
NavComposer formulates instruction generation as an egocentric video captioning task and employs a two-stage pipeline: entity extraction and instruction synthesis.
In the entity extraction, visual observations are decomposed into three types of semantic entities: action (e.g., ``turn left"), scene (e.g., ``hallway"), and object (e.g., ``table").
In the instruction synthesis, it then recompose the list of entities and synthesizes concise, natural-language instructions.
The modular architecture of NavComposer offers high flexibility.
Each module is independently configured, allowing seamless integration of both classical techniques like visual odometry, and advanced models like LLMs.
By explicitly extracting semantic entities, NavComposer captures the essential components of trajectories, significantly improving instruction quality.
This entity-based modularization also enhances control over the generation process, enabling style customization and enriching linguistic diversity.
Furthermore, NavComposer is dataset-agnostic, eliminating dependence on specific datasets or training regimens.
It can process general egocentric videos regardless of recording devices or content domains.
This adaptability makes NavComposer suitable for both large-scale internet datasets and real-world robotic applications.
Experiments demonstrate that our modular framework significantly outperforms the end-to-end trained baseline generator.
Extensive results further show that NavComposer can be adapted to 10 diverse sources spanning various devices, domains, and scales.

We also introduce NavInstrCritic, a comprehensive, annotation-free evaluation system for navigation instructions.
Existing evaluation protocols primarily assess how generated instructions indirectly improve navigation models \cite{magassouba2021crossmap} or mimic a limited set of human annotations \cite{wu2021improved}.
In contrast, NavInstrCritic is decoupled from navigation agents and directly evaluates instructions.
Recognizing the multifaceted nature of trajectory descriptions, NavInstrCritic assesses instructions across three dimensions: contrastive matching (overall alignment with trajectories), semantic consistency (detailed correspondence with semantic entities), and linguistic diversity (richness and variability of language).
By eliminating reliance on expert annotations and incorporating underexplored diversity metrics, NavInstrCritic offers a generalizable and holistic evaluation framework applicable to various domains.

Our key contributions are as follows:

\begin{itemize}
    \item We propose NavComposer, a novel modular framework for navigation instruction generation. By explicitly modeling actions, scenes, and objects as semantic entities and leveraging LLMs, NavComposer enhances flexibility, quality, and adaptability of instruction generation.
    \item We introduce NavInstrCritic, a comprehensive and annotation-free evaluation system that assesses navigation instructions across three dimensions, offering a holistic and generalizable instruction analysis.
    \item Extensive experiments demonstrate the effectiveness of our method, showing improvements over existing models and highlighting its potential for broad applications.
\end{itemize}

\section{Related Work}
\subsection{Language-guided Navigation}
Language-guided navigation is a cornerstone of embodied AI, enabling agents to interpret and follow natural language instructions in complex environments.
A key task in this domain is Vision-and-Language Navigation (VLN) \cite{anderson8578485}, where agents rely solely on visual observations to execute language-based navigation.
The introduction of datasets like R2R \cite{anderson8578485}, RxR \cite{ku2020room}, and Touchdown \cite{chen8954308} has significantly advanced VLN research by providing standardized benchmarks.
Existing methods predominantly use deep neural networks, including sequence-to-sequence models \cite{zhu2020vision, zhang2021spc-nav, zhang2021language}, feature-enhanced techniques \cite{wang2023dual, Wang_2024_CVPR}, waypoint-based approaches \cite{krantz2021waypoint, hong2022bridging, he2024instruction}, and transformer-based architectures \cite{majumdar2020improving, hong2021vln, chen2022think, he2024multimodal}.
Despite these advances, training data remains scarce due to the high cost of expert annotations, limiting dataset scalability and generalizability.
To address this challenge, we propose NavComposer, a novel framework that automatically composes language instructions for any given navigation trajectory.
By generating large-scale high-quality instructions, NavComposer mitigates data scarcity issues and enables scalable, adaptable language-guided navigation.


\subsection{Instruction Augmentation Models}
Instruction augmentation methods, commonly known as speaker models, have recently played a critical role in VLN.
Early models \cite{fried2018speaker, tan2019learning, agarwal2019visual} used an LSTM-based sequence-to-sequence framework to generate navigation instructions.
For example, EnvDrop \cite{tan2019learning} improved the genralization of instructions by randomly dropping out environment features.
However, these models often produce repetitive or inconsistent outputs due to limited training data and capacities.
Recent approaches integrate transformer models \cite{magassouba2021crossmap, wu2021improved, Wang_2023_CVPR, wang2024pasts} and diverse trajectory data \cite{li2022envedit, wang2022less, lin2023learning}.
LANA \cite{Wang_2023_CVPR}, for instance, presents an impressive solution by jointly learning instruction following and generation within a single model.
To our knowledge, these advancements are tightly coupled to specific datasets and agents, with no prior work addressing a general instruction generator.
In our work, NavComposer introduces a modular framework that explicitly models semantic entities, thereby improving the semantic consistency of generated instructions.
Thanks to its modular and dataset-agnostic design, NavComposer can process diverse RGB navigation video without requiring additional training, enabling broader adaptability.

\subsection{Video Captioning}
Video captioning (VC) \cite{MOCTEZUMA2023103671, abdar2024review, sanders2024survey} shares similarities with navigation instruction generation, as both translate visual inputs into natural language.
Recent advancements in video captioning include a dual-stream RNN \cite{xu2019dsrnn} for jointly discovering and integrating visual and semantic streams, global-local feature representations via multiple encoders at different granularities \cite{yan2022glr}, and element-aware frameworks incorporating two element-specific linguistic features \cite{liu2024evcap}.
More recently, multimodal LLMs \cite{wang2025internvideo2, Qwen2-VL} have furthered video understanding.
Despite these insights, directly transferring VC methods to the navigation domain is challenging
One key difference is that video captioning primarily describes observed events (e.g., “A person is folding paper on a table”), whereas navigation instructions are egocentric and action-driven (e.g., “Turn left until you can’t see the table”).
Additionally, common VC datasets, such as MSR-VTT \cite{xu2016msr} and WebVid-2M \cite{bain2021frozen}, focus on external events on third-person perspectives rather than egocentric motion.
To address this gap, we propose a specialized approach that explicitly extracts semantic entities for navigation-relevant instructions.
Our work also highlights an underexplored direction: automatically generating text annotations for egocentric videos.

\begin{figure*}[tbp]
    \centering
    \includegraphics[width=0.98\textwidth]{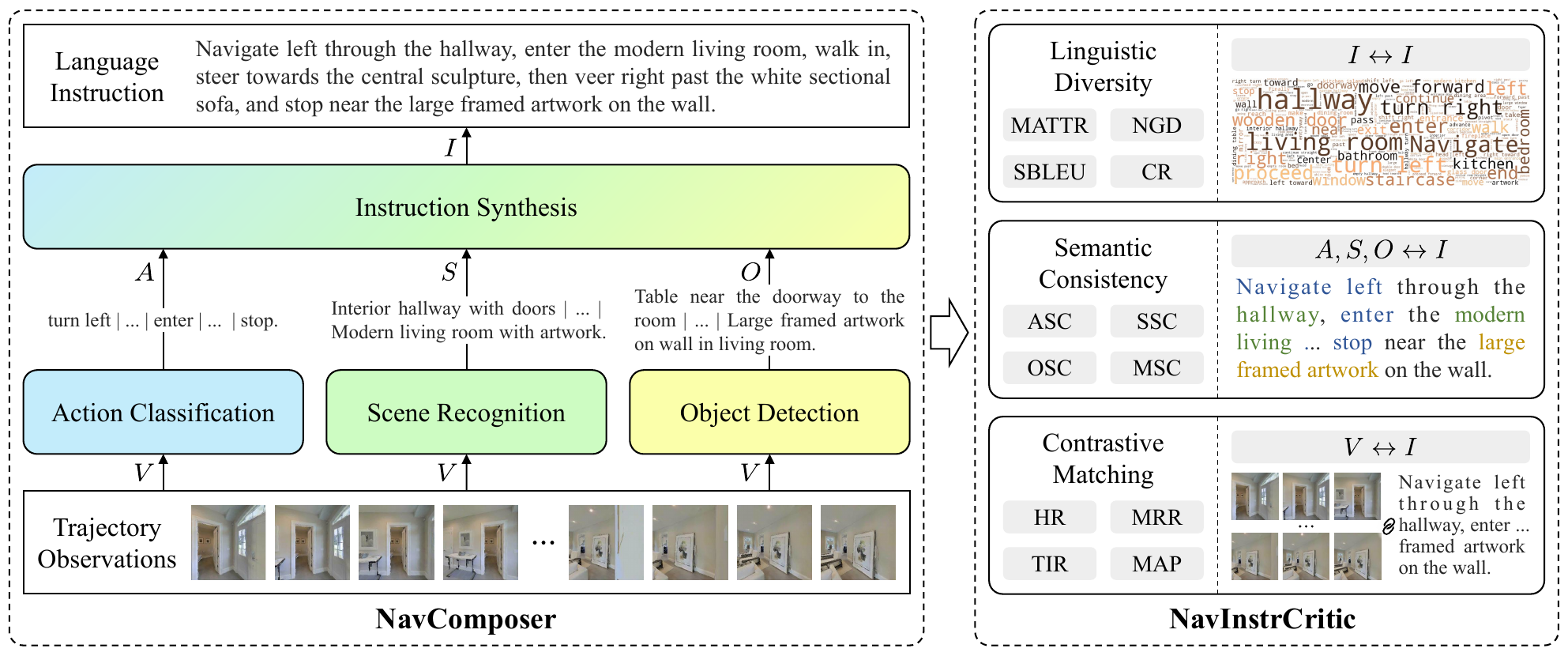}
    \caption{NavComposer processes visual observations from a navigation trajectory, extracts semantic entities, and synthesizes a fluent language instruction, where $V$, $A$, $S$, $O$, and $I$ denote visual observations, actions, scenes, objects, and generated instructions, respectively. NavInstrCritic then evaluates instruction quality across three dimensions. Each dimension involves specific alignments (e.g., $V \leftrightarrow I$ denotes the alignment evaluation between visual observations and instructions) and corresponding metrics (e.g., HR refers to hit rate), as detailed in Section~\ref{sec:method_navinstrcritic}.}
    \label{fig:method_architecture}
\end{figure*}

\section{Methodology}
As illustrated in Fig.~\ref{fig:method_architecture}, the generation of navigation instructions is formulated as video-to-text translation.  
Given a sequence of visual observations from a navigation trajectory, $V = \{v_1, \dots, v_T\}$, NavComposer first extracts key semantic entities.  
Specifically, three independent modules extract action entities $A = \{a_1, \dots, a_T\}$, scene entities $S = \{s_1, \dots, s_T\}$, and object entities $O = \{o_1, \dots, o_T\}$, where each captures distinct information from its corresponding frames.  
Then, the instruction synthesis module integrates $A$, $S$, and $O$ to generate a fluent language instruction $I$.  
Afterward, the complete set of information—$V$, $A$, $S$, $O$, and $I$—undergoes a comprehensive evaluation using the NavInstrCritic system, which provides a multifaceted assessment of instruction quality.
In this work, the terms image sequence, video, and visual observations from the navigation trajectory are used interchangeably.

\subsection{NavComposer Framework}
The proposed NavComposer framework generates navigation instructions by decomposing semantic entities from the input video and recomposing them into coherent natural-language instructions.
It consists of four core modules: action classification, scene recognition, object detection, and instruction synthesis.
The action classification module identifies navigation-related actions.
The scene recognition module perceives and describes the surroundings.
The object detection module identifies significant landmarks.
Finally, the instruction synthesis module leverages language as a bridge to integrate semantic entities into a natural language instruction.

\subsubsection{Action Classification Module}
The action classification module $f_{\text{action}}$ analyzes the relative motion between consecutive frames in the input video $V = {v_1, \dots, v_T}$.
Specifically, it predicts the robot’s action $a_t$ based on the visual input from adjacent frames $(v_t, v_{t+1})$:
\begin{equation}
    a_t = 
    \begin{cases} 
        f_{\text{action}}(v_t, v_{t+1}), & \text{if } t < T, \\
        \text{``stop"}, & \text{if } t = T.
    \end{cases}
\end{equation}
Similar to VLN-CE \cite{krantz2020beyond}, we use the action space $a_t \in\ $\{``stop", ``move forward",``turn left",``turn right"\}. 
For the last frame ($t = T$), the action is manually set to ``stop", indicating the end of the trajectory.

\begin{figure*}[tbp]
\centering
    \subfloat[Action]{\includegraphics[width = 0.32\textwidth]{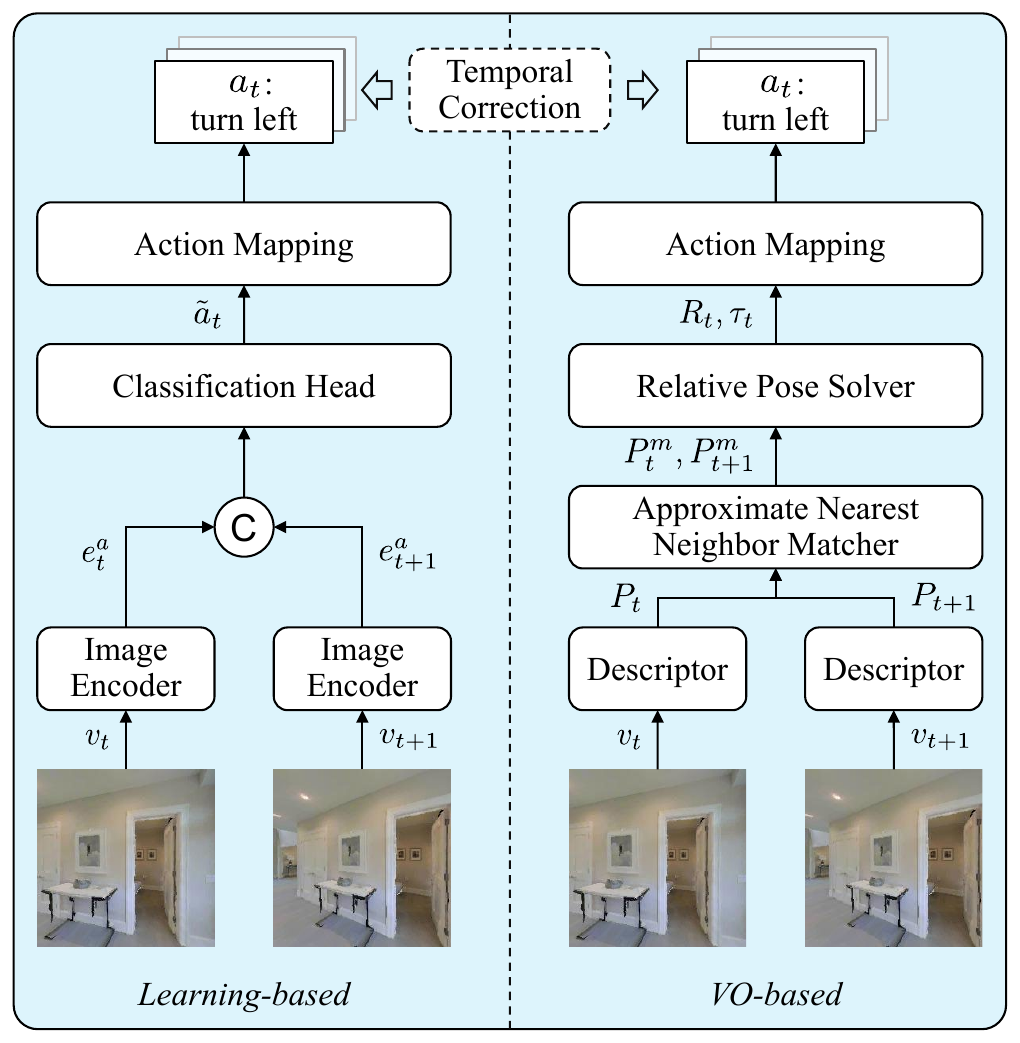}\label{fig:module_action}}
    \hfill
    \subfloat[Scene]{\includegraphics[width = 0.32\textwidth]{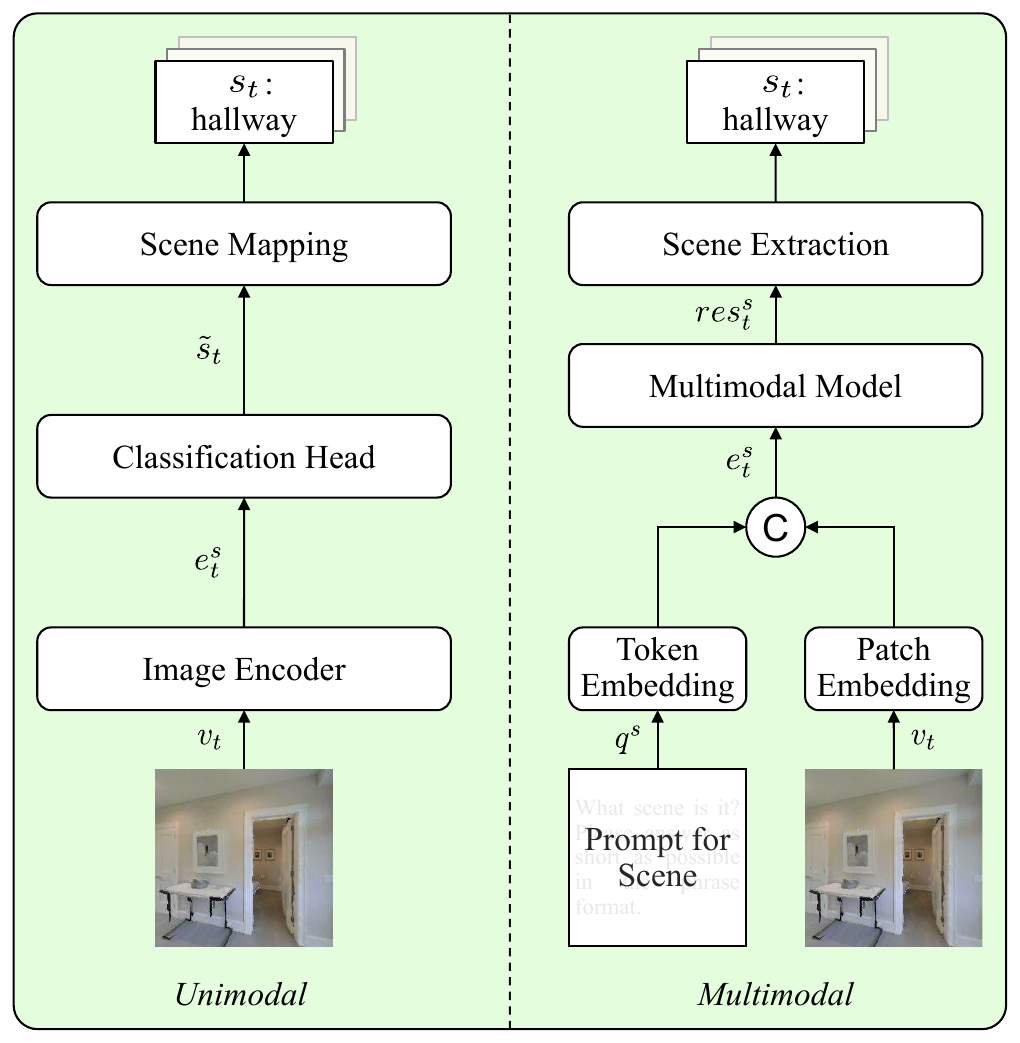}\label{fig:module_scene}}
    \hfill
    \subfloat[Object]{\includegraphics[width = 0.32\textwidth]{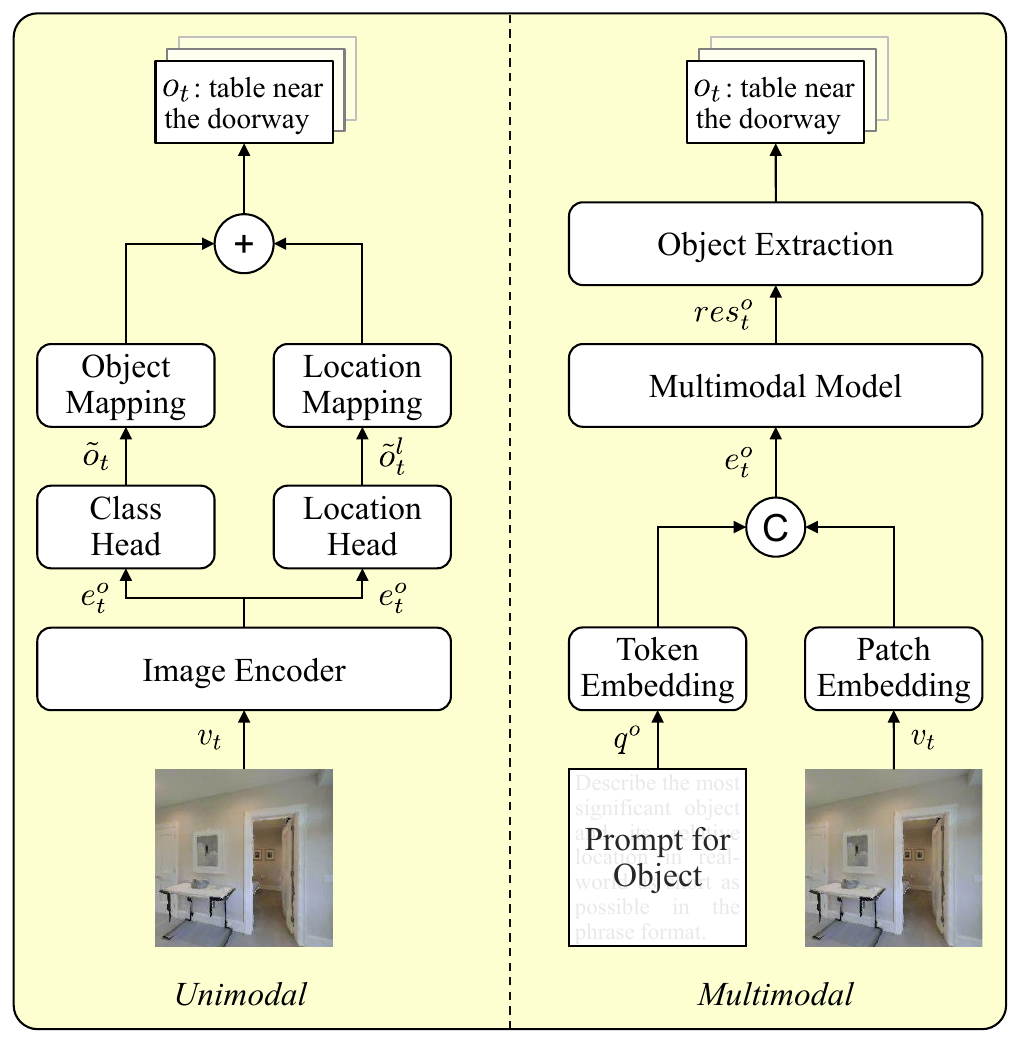}\label{fig:module_object}} 
\caption{Details of the action classification, scene recognition, and object detection modules. Each module supports two types of architectures with various implementations.}
\label{fig:module_detail}
\end{figure*}

To implement $f_{\text{action}}$, we explore two categories: learning-based and visual odometry-based, as illustrated in Fig.~\ref{fig:module_action}.
In the learning-based approach, we design a lightweight action classifier that leverages a pre-trained image encoder, such as ResNet \cite{he2016deep}, with a trainable fully connected layer (FC) to predict actions:
\begin{align}
    e^a_t &= \text{Encoder}(v_t), \\
    e^a_{t+1} &= \text{Encoder}(v_{t+1}), \\
    \tilde{a}_t &= \text{FC}([e^a_t; e^a_{t+1}]), \\
    a_t &= \text{Argmax}(\tilde{a}_t),
\end{align}
where $\text{Encoder}()$ represents a pre-trained image encoder, $[\cdot; \cdot]$ denotes feature concatenation, and $\tilde{a}_t$ is the predicted distribution.

In the visual odometry-based approach, we estimate relative motion between frames using visual odometry techniques:
\begin{align}
    P_t, D_t & = \text{SIFT}(v_t), \\ 
    P_{t+1}, D_{t+1} & = \text{SIFT}(v_{t+1}), \\
    P^m_t, P^m_{t+1} & = \text{FLANN}(P_t, D_t, P_{t+1}, D_{t+1}), \\
    R, \tau & = \text{RecoverPose}(P^m_t, P^m_{t+1}, K), \\
    a_t & = \text{ClassifyPose}(R, \tau),
\end{align}
where $\text{SIFT}()$ extracts keypoints and descriptors \cite{sift790410}, $\text{FLANN}()$ matches keypoints between frames \cite{flann}, $\text{RecoverPose}()$ estimates the relative rotation $R$ and translation $\tau$, $K$ is intrinsic parameters, and $\text{ClassifyPose}()$ maps these transformations to predefined action classes.

Action prediction from consecutive RGB images is inherently challenging and prone to errors. 
To improve sequence consistency, we introduce two temporal correction principles:
1) Sudden oscillations between actions (e.g., ``move forward, turn right, move forward”) are unlikely or just slight adjustments.
When an $\textit{ABA}$ pattern is detected, it is corrected to $\textit{AAA}$.
2) Turning actions that cancel each other (e.g., ``turn left, turn left, turn right”) are invalid in the shortest path.
Patterns like $\textit{AAB}$ are corrected to $\textit{AAA}$ using majority voting.
These mechanisms enhance the smoothness and reliability of the predicted action sequence, mitigating potential errors.

\subsubsection{Scene Recognition Module}
In indoor navigation tasks, scene description can enhance robot's spatial cognition and inspire its commonsense knowledge. 
Thus, the scene recognition module $f_{\text{scene}}$ provides an overall understanding of the spatial environment:
\begin{equation}
    s_t = f_{\text{scene}}(v_t),
\end{equation}
where $s_t$ is a phrase describing the current scene.

There are two main approaches to implementing the scene recognition module: unimodal methods and multimodal methods, as shown in Fig. \ref{fig:module_scene}.
Unimodal methods, such as SWAG \cite{Singh_2022_CVPR} only accept images as input:
\begin{align}
    e^s_t &= \text{Encoder}(v_t), \\
    \tilde{s}_t &= \text{FC}(e^s_t), \\
    s_t &= \text{Argmax}(\tilde{s}_t),
\end{align}
where $e^s_t$ represents the encoded features of the input image, $\tilde{s}_t$ is the classification distribution, and $s_t$ is the predicted scene category.
However, unimodal methods are typically designed for specific datasets and may struggle to generalize to categories outside their predefined domains. 

An alternative approach leverages multimodal LLMs, such as BLIP2 \cite{pmlr-v202-li23q}, which benefit from large-scale pretraining. 
These models accepts both image and text as inputs, and are applicable to various domains without requiring domain-specific training:
\begin{align}
    e^s_t &= [\text{Emb}(q^s), \text{Emb}(v_t)], \\
    res^s_t &= \text{MultimodalLLM}(e^s_t), \\
    s_t &= \text{Extract}(res^s_t),
\end{align}
where $q^s$ is the query prompt, $\text{Emb}()$ denotes embedding functions, $res^s_t$ is the model's response, and $s_t$ is the extracted scene description.


\subsubsection{Object Detection Module}
The object detection module $f_{\text{object}}$ identifies landmarks that contribute to the navigation trajectory:
\begin{equation}
    o_t = f_{\text{object}}(v_t).
\end{equation}

To implement this module, we incorporate both unimodal and multimodal techniques, as shown in Fig.~\ref{fig:module_object}.  
Unimodal approaches such as DETR \cite{nicolas2020detr} are widely used for object detection.  
The most significant object is selected based on size or confidence score.  
For object localization, our module emphasizes real-world spatial relationships (e.g., ``trees far right," ``painting on wall") rather than image coordinates.  
To encode such relationships, the image is divided into a nine-square grid, and the detected bounding box is mapped to a predefined spatial description.  
This rule leverages the egocentric nature of input images and, though not highly precise, aligns with common linguistic patterns.  
The process can be summarized as follows:
\begin{align}
    e^o_t & = \text{Encoder}(v_t), \\
    \tilde{o}_t & = \text{FC}(e^o_t), \tilde{o}^l_t= \text{FC}^l(e^o_t), \\
    o_t & = \text{Aggregate}(\tilde{o}_t, \tilde{o}^l_t),
\end{align}
where $e^o_t$ represents the encoded features of the input image, $\tilde{o}_t$ denotes the detected object categories, $\tilde{o}^l_t$ captures the spatial relationships, and $o_t$ aggregates these components into a comprehensive object description.

Despite their effectiveness, traditional models are typically trained to optimize accuracy on specific datasets and detect only a predefined set of objects.
To address this limitation, we leverage multimodal LLMs as open-ended visual question-answering systems.
Pretrained on large-scale data, models like LLaVA \cite{liu2023visual} possess extensive commonsense knowledge and can recognize a diverse range of objects.  
Furthermore, these models infer spatial relationships between objects without the need for manual mapping rules.
The process involving multimodal LLMs is as follows:
\begin{align}
    e^o_t & = [\text{Emb}(q^o),\text{Emb}(v_t)] \\
    res^o_t & = \text{MultimodalLLM}(e^o_t), \\
    o_t & = \text{Extract}(res^o_t),
\end{align}
where $q^o$ is the query prompt, $\text{Emb}()$ denotes embedding functions, $res^o_t$ is the model's response, and $o_t$ is the extracted object description.

\subsubsection{Instruction Synthesis Module}
Language acts as a bridge in the instruction synthesis module, $I = f_{\text{summary}}(A,S,O)$, by integrating navigation actions, traversed scenes, and key objects to produce precise and diverse navigation instructions.  
Algorithm~\ref{alg:module_synthesis} presents a concise pseudocode.

We introduce action-based key frame downsampling to tackle the potential redundancy in semantic entities.
The start, middle, and end frames of consecutive identical actions are defined as key frames. 
The selection of a key frame is determined by both the action type and a uniform random variable.
Specifically, for actions other than ``move forward,” the middle key frame is always selected.
For ``move forward” actions, there is a $1/6$ probability of being treated as ``enter", ``pass", or ``leave," corresponding to the use of the start, middle, or end frame, respectively.
This mechanism enriches action descriptions and reduces computational costs without compromising essential navigation information.

To introduce linguistic diversity, we propose a double random synonym replacement process.
Let $E$ denote the synonym set for each element $e \in {a, s, o}$.
First, we sample a ratio $M \sim\text{Uniform}(0,1), M \in [0,1]$. 
This occurs once pre-generation, keeping $M$ consistent across all steps.
Let $N$ denote the size of $E$. 
Next, we randomly select a synonym index $X$ from a subset of $E$: $X \sim\text{UniformInt}(1, \lceil MN \rceil), X\in \{1,2,\dots,\lceil MN \rceil\}$.
The final selected synonym is the $X$-th element in $E$. 
This process enhances linguistic diversity and can simulate language preferences by structuring synonym sets.

\begin{algorithm}[tbp]
\caption{Instruction Synthesis}
\label{alg:module_synthesis}
\KwIn{Multimodal entities $A, S, O$, Prompt $prom$}
\KwOut{Language instruction $I$}

\tcp{Key frame downsampling}
$A^c, S^c, O^c \gets \emptyset$\; 
$i \gets 1$\;
\While{$i \leq T$}{
    $j \gets i + 1$\;
    \While{$j \leq T$ \textbf{and} $a_i = a_j$}{
        $j \gets j + 1$\; 
    }
    
    Sample $Z$ from $\text{Uniform}(0, 1)$\;
    \If{$a_i=\texttt{"move forward"}$ \textbf{and} $Z\le 1/2$}{
        \If{$Z\le 1/6$}{
            $k = i$,
            $a_{k}\gets \texttt{"enter"}$\; 
        }\ElseIf{$Z\le 1/3$}{
            $k = \lfloor (i + j - 1) / 2 \rfloor$,
            $a_{k}\gets \texttt{"pass"}$\; 
        }\Else{
            $k = j-1$,
            $a_{k}\gets \texttt{"leave"}$\; 
        }
    }\Else{
        $k = \lfloor (i + j - 1) / 2 \rfloor$;
    }

    Add $a_{k}, s_{k}, o_{k}$ into $A^c, S^c, O^c$\; 
    $i \gets j$\; 
}

\tcp{Random replacement}
$A^r, S^r, O^r \gets \emptyset$\; 
Sample $M$ from $\text{Uniform}(0, 1)$\; 
\For{$i \gets 1$ \textbf{to} $T$}{
    $E^a, E^s, E^o \gets \text{PossibleReplacements}(a^c_i, s^c_i, o^c_i)$\; 
    $N^a, N^s, N^o \gets |E^a|, |E^s|, |E^o|$\; 
    $N^{ar}, N^{sr}, N^{or} \gets \lceil M \cdot N^a \rceil, \lceil M \cdot N^s \rceil, \lceil M \cdot N^o \rceil$\; 
    Sample $X^a, X^s, X^o$ from $\text{UniformInt}(1, N^{ar})$, $\text{UniformInt}(1, N^{sr})$, $\text{UniformInt}(1, N^{or})$\; 
    $a^r_i, s^r_i, o^r_i \gets E^a_{X^a}, E^s_{X^s}, E^o_{X^o}$\; 
}

\tcp{LLM-based synthesis}
$desc \gets \texttt{""}$\; 
\For{$i \gets 1$ \textbf{to} $T$}{
    $curr \gets \text{OrganizeEntity}(a^r_i, s^r_i, o^r_i)$\; 
    $desc \gets \text{OrganizeList}(desc, curr)$\; 
}
$I \gets \text{LLM}(prom, desc)$\; 
\Return $I$\; 
\end{algorithm}

Finally, entity sequences are temporally ordered for clear step-by-step navigation descriptions.
$a_t$, $s_t$, and $o_t$ are concatenated via $OrganizeEntity()$ in temporal order with predefined separators.
$A$, $S$, and $O$ are merged using $OrganizeList()$ as a list string.
We then design a tailored prompt to guide the LLM in generating accurate, context-aware instructions.
LLMs offer key advantages: they effectively bridge multimodal information through language, require no additional training due to their rich linguistic knowledge, and allow flexible control over output via prompt modification.
After concatenating the prompt with the temporal descriptions, the LLM processes the input to produce a navigation instruction $I$.

\subsection{NavInstrCritic System}
\label{sec:method_navinstrcritic}
In language-guided navigation, instruction evaluation typically relies on translation metrics (e.g., BLEU \cite{papineni2002bleu}) using dataset annotations as ground truth. 
However, navigation trajectories allow diverse valid descriptions, and reliance on limited expert annotations introduces bias. 
Evaluating via navigation performance after augmentation is also suboptimal, as confounding factors may affect the agent.

To address these issues, we propose NavInstrCritic, a novel and comprehensive evaluation system for assessing navigation language instructions.  
NavInstrCritic evaluates instructions across three main dimensions: contrastive matching, semantic consistency, and linguistic diversity, as shown in Fig.~\ref{fig:method_architecture}.  
Each dimension consists of four metrics, providing a holistic assessment of instruction quality.  
Additionally, given the lack of prior work on instruction generation for continuous domain, we introduce a new end-to-end baseline model alongside this evaluation system.

\subsubsection{Contrastive Matching}
We formulate the alignment between a navigation trajectory and its corresponding instruction as a contrastive matching problem ($V\leftrightarrow I$).  
To achieve this, we train a contrastive matcher, $g_{\text{cm}}$, that maps both the navigation trajectory and the instruction into a unified feature space and computes their cosine similarity.
The matcher is based on CLIP \cite{pmlr-v139-radford21a-clip}, with modifications to the pooling layer to accommodate image sequence inputs.
After training, the matcher can assign similarity scores to each trajectory-instruction pair.
For a batch of size $B$, the similarity between the $p$-th trajectory $V^{(p)}$ and the $q$-th generated instruction $I^{(q)}$ is computed as:
\begin{equation} 
Sim_{p,q} = \text{CosineSim}(g_{\text{cm}}(V^{(p)}), g_{\text{cm}}(I^{(q)})). 
\end{equation}
Using the similarity matrix $Sim$ within a batch as ranking evidence, we compute several common retrieval metrics:
\begin{itemize}
    \item \textbf{Hit Rate (HR)}: Measures the proportion of relevant items found within the top-$k$ results.  
    \item \textbf{Mean Reciprocal Rank (MRR)}: Computes the average reciprocal rank of the first relevant item for each query, emphasizing the position of the first match.  
    \item \textbf{Trajectory-to-Instruction Recall (TIR)}: Evaluates retrieval performance by computing the ratio of retrieved relevant items to the total number of relevant items.  
    \item \textbf{Mean Average Precision (MAP)}: Averages precision across all ranks for each query, balancing precision and recall.  
\end{itemize}

\subsubsection{Semantic Consistency}
Beyond overall contrastive matching, we evaluate the semantic consistency of generated instructions ($A,S,O\leftrightarrow I$).  
Using LLMs, we assess whether the instructions accurately reflect key entities in the navigation trajectory, including actions, scenes, and objects.  
This consistency is quantified between 0-10 using three distinct scores and an overall mean metric:
\begin{itemize} 
    \item \textbf{Action Semantic Consistency (ASC)}: Measures the alignment of navigation actions.  
    \item \textbf{Scene Semantic Consistency (SSC)}: Evaluates the consistency of traversed scenes.  
    \item \textbf{Object Semantic Consistency (OSC)}: Assesses the alignment of observed objects.  
    \item \textbf{Mean Semantic Consistency (MSC)}: Computes the average of ASC, SSC, and OSC, providing an overall measure of semantic consistency.  
\end{itemize}

\subsubsection{Linguistic Diversity}
Linguistic diversity captures variations in structure, vocabulary, and expression while conveying the same idea.  
While accuracy is paramount, enhancing diversity can improve generalization in navigation models.  
Despite its importance, this aspect has received little attention in prior research.  
To quantify linguistic diversity ($I\leftrightarrow I$), we suggest the following four metrics:
\begin{itemize} 
    \item \textbf{Moving Average Token-Type Ratio (MATTR)} \cite{Covington01052010}: Measures the proportion of unique token types relative to the total number of tokens within a sliding window, assessing vocabulary variation across different text segments.  
    \item \textbf{N-gram Diversity (NGD)} \cite{10.1162/tacl_a_00536_ngd}: Evaluates variation in n-grams within the generated text. Higher NGD scores indicate greater diversity and reduced repetition.  
    \item \textbf{Self-BLEU (SBLEU)} \cite{zhu2018selfbleu}: Computes BLEU scores between different sentences in a corpus. Lower SBLEU values indicate greater diversity and reduced redundancy.
    \item \textbf{Compression Ratio (CR)} \cite{shaib2024standardizing}: Computes the ratio of the original text length to its compressed length. Lower CR suggests greater diversity.
\end{itemize}


\begin{table}[tbp]
\centering
\caption{Module implementations and corresponding abbreviations.}
\label{tab:module_implementations}
\begin{tabular}{llL}
\toprule
Module                   & Implementation                                    & \textrm{Abbr.} \\ \midrule
\multirow{3}{*}{Action}  & ResNet50 \cite{he2016deep} finetuned on VLN-CE \cite{krantz2020beyond}          & rn           \\
                         & DINOv2-B/14 \cite{oquab2023dinov2} finetuned on VLN-CE \cite{krantz2020beyond}        & dn           \\
                         & SIFT \cite{sift790410} + FLANN \cite{flann}                 & vo           \\ \midrule
\multirow{6}{*}{Scene}   & MAE-B/16 \cite{He_2022_CVPR} finetuned on Places365 \cite{7968387places365}                & mae          \\
                         & SWAG-L/16 \cite{Singh_2022_CVPR}                & swg          \\
                         & BLIP-2 \cite{pmlr-v202-li23q}                       & blp          \\
                         & LLaVA-1.6 \cite{liu2024llavanext}                   & llv          \\
                         & Qwen2.5-VL-7B-Instruct \cite{Qwen2.5-VL}                            & qwn          \\
                         & GPT-4o-mini  \cite{gpt-4o-mini}                                & gpt          \\ \midrule
\multirow{6}{*}{Object}  & DETR \cite{nicolas2020detr}                             & dtr          \\
                         & SWAG-L/16 \cite{Singh_2022_CVPR}                & swg          \\
                         & BLIP-2 \cite{pmlr-v202-li23q}                          & blp          \\
                         & LLaVA-1.6 \cite{liu2024llavanext}                     & llv          \\
                         & Qwen2.5-VL-7B-Instruct   \cite{Qwen2.5-VL}                            & qwn          \\
                         & GPT-4o-mini \cite{gpt-4o-mini}                                   & gpt          \\ \midrule
\multirow{4}{*}{Synthesis} & Llama-3.1-8B-Instruct  \cite{grattafiori2024llama}                          & llm          \\
                         & Gemma-2-9b-it   \cite{team2024gemma}                                 & gmm          \\
                         & Qwen2.5-7B-Instruct  \cite{qwen2.5}                             & qwn          \\
                         & GPT-4o-mini \cite{gpt-4o-mini}                                    & gpt          \\ \bottomrule
\end{tabular}
\end{table}

\section{Experiments}
\subsection{Experimental Setup}
\label{sec:exp_setup}
\textbf{NavComposer-related.} We evaluate several module implementations within NavComposer, including odometry-based, learning-based, and multimodal LLM approaches.
Table~\ref{tab:module_implementations} summarizes the implementations, corresponding versions, and respective abbreviations.
To systematically name each variant, we adopt a straightforward convention that concatenates the abbreviations of individual modules.
For instance, \texttt{vo-llv-llv-llm} denotes the use of visual odometry (\texttt{vo}) for the action module, LLaVA (\texttt{llv}) for both the scene and object modules, and Llama (\texttt{llm}) for instruction synthesis.
To ensure a fair comparison, all offline LLMs have approximately 7B parameters.
We analyze the performance differences among these implementations in Section~\ref{sec:exp_flexibility}.
Based on this analysis, we select the best-performing variant for further comparisons in instruction quality (Section~\ref{sec:exp_quality}) and adaptability (Section~\ref{sec:exp_adaptability}).

\textbf{NavInstrCritic-related.} The proposed NavInstrCritic system evaluates the quality of generated navigation instructions across three key dimensions: contrastive matching, semantic consistency, and linguistic diversity.
For contrastive matching, we train a matcher model using pretrained CLIP weights from OpenCLIP \cite{ilharco_gabriel_2021_5143773}. 
This model maps both trajectories and instructions into a 768-dimensional feature space, enabling cosine similarity computation. 
We then compute retrieval metrics using TorchMetrics \cite{Detlefsen2022torchmetrics}.
For semantic consistency, we leverage the Qwen-14B large language model \cite{qwen2.5} to assess alignment with key navigation entities—actions, scenes, and objects. 
We extract consistency scores from the model’s responses using pattern matching. 
For methods that do not explicitly extract semantic entities, we use unified results from our best-performing variant.
For linguistic diversity, we compute various metrics using lex\_div, NLTK, and gzip tools. 
Metrics SBLEU and CR are denoted with $\downarrow$ (lower is better), while higher values indicate better performance for all other metrics.

\textbf{End-to-end Baseline.} To the best of our knowledge, no prior work has specifically addressed navigation instruction generation in continuous domains \cite{krantz2020beyond}. 
To complement our evaluation system, we introduce an end-to-end baseline generator, which is included in the comparisons presented in Section~\ref{sec:exp_quality}. 
This model adopts a sequence-to-sequence architecture trained directly on the target dataset. 
Its backbone is CoCa \cite{yu2022coca}, a vision-to-text model designed for contrastive learning and generation. 
To adapt CoCa for continuous navigation trajectories, we modify the pooling layers of its vision encoder to handle video input. 
This baseline is fine-tuned on the VLN-CE training dataset \cite{krantz2020beyond}, enabling end-to-end instruction generation for continuous VLN tasks. 
Additional details are provided in Appendix B.

\begin{table*}[tbp]
\centering
\caption{Performance of different action classification module implementations. The last column reports action accuracy as a percentage. \textbf{Bold} indicates the best performance, while \underline{underlined} indicates the second-best.}
\label{tab:module_action}
\begin{tabular}{Lccccccccccccc}
\toprule
   \multirowcenter{2}{*}{\textrm{Variant}} & \multicolumn{4}{c}{Contrastive Matching} & \multicolumn{4}{c}{Semantic Consistency} & \multicolumn{4}{c}{Linguistic Diversity} & \multirowcenter{2}{*}{\makecell{Action \\ Accuracy (\%)}} \\ \cmidrule(lr){2-5}\cmidrule(lr){6-9}\cmidrule(l){10-13}
   & HR       & MRR      & TIR      & MAP     & ASC     & SSC     & OSC     & MSC     & MATTR      & NGD     & SBLEU$\downarrow$   & CR$\downarrow$      &    \\ \midrule
vo-llv-llv-llm & \underline{0.731} &  \underline{0.615} &  \underline{0.489} &  \underline{0.607} &  \underline{3.882} &  \underline{2.999} &  \underline{4.140} &  \underline{3.674} &  \underline{0.580} &  \underline{0.973} &  \underline{0.893}   &  \underline{5.379} & \textbf{98.5}   \\
dn-llv-llv-llm & \textbf{0.740} &  \textbf{0.626} &  \textbf{0.502} &  \textbf{0.618} &  \textbf{4.006} &  2.937 &  4.076 &  3.673 &  0.579 &  0.938 &  0.898   &  5.410 & \underline{85.1}   \\
rn-llv-llv-llm & 0.682 &  0.542 &  0.430 &  0.539 &  3.181 &  \textbf{3.578} &  \textbf{5.748} &  \textbf{4.169} &  \textbf{0.606} &  \textbf{1.096} &  \textbf{0.876}   &  \textbf{5.260} & 67.3   \\ \bottomrule
\end{tabular}
\end{table*}

\begin{table*}[tbp]
\centering
\caption{Performance of different scene recognition module implementations. Scene Length column represents the average number of words in a scene description. \textbf{Bold} indicates the best performance, while \underline{underlined} indicates the second-best.}
\label{tab:module_scene}
\begin{tabular}{Lccccccccccccc}
\toprule
   \multirowcenter{2}{*}{\textrm{Variant}} & \multicolumn{4}{c}{Contrastive Matching} & \multicolumn{4}{c}{Semantic Consistency} & \multicolumn{4}{c}{Linguistic Diversity} & \multirowcenter{2}{*}{\makecell{Scene \\ Length}} \\ \cmidrule(lr){2-5}\cmidrule(lr){6-9}\cmidrule(l){10-13}
   & HR       & MRR      & TIR      & MAP     & ASC     & SSC     & OSC     & MSC     & MATTR      & NGD     & SBLEU$\downarrow$   & CR$\downarrow$      &    \\ \midrule
vo-llv-llv-llm & 0.731 &  0.615 &  0.489 &  0.607 &  \underline{3.882} &  2.999 &  4.140 &  3.674 &  0.580 &  0.973 &  0.893   &  5.379 & 1.487   \\
vo-mae-llv-llm & 0.745 &  0.621 &  0.469 &  0.613 &  3.555 &  3.080 &  3.744 &  3.460 &  \underline{0.593} &  \textbf{1.130} &  \underline{0.866}   &  5.108 & 1.507   \\
vo-swg-llv-llm & 0.733 &  0.611 &  0.444 &  0.601 &  3.550 &  2.633 &  \textbf{4.219} &  3.468 &  0.589 &  \underline{1.101} &  0.868   &  \underline{5.086} & 1.332   \\
vo-blp-llv-llm & \underline{0.781} &  \underline{0.654} &  0.500 &  \underline{0.647} &  3.810 &  \underline{3.347} &  4.043 &  \underline{3.734} &  0.575 &  0.928 &  0.891   &  5.364 & 7.807   \\
vo-qwn-llv-llm & 0.758 &  0.637 &  \underline{0.506} &  0.630 &  3.692 &  3.062 &  4.106 &  3.620 &  0.588 &  0.976 &  0.895   &  5.329 & 1.278   \\
vo-gpt-llv-llm & \textbf{0.808} &  \textbf{0.672} &  \textbf{0.521} &  \textbf{0.664} &  \textbf{4.269} &  \textbf{4.136} &  \underline{4.177} &  \textbf{4.194} &  \textbf{0.599} &  1.074 &  \textbf{0.864}   &  \textbf{5.063} & 4.098   \\ \bottomrule
\end{tabular}
\end{table*}

\textbf{Datasets.} The validation data for Sections~\ref{sec:exp_flexibility} and \ref{sec:exp_quality} is sourced from VLN-CE \cite{krantz2020beyond}. 
We combine the seen and unseen splits, yielding 872 navigation trajectories. 
To mitigate random variation, we sample three instructions per trajectory, resulting in 2,616 instructions.
In Section~\ref{sec:exp_adaptability}, we further incorporate 10 diverse datasets to evaluate NavComposer’s adaptability, generating totally 232,569 instructions (see Table~\ref{tab:diverse_generation}).
More information on LLM prompts and module training can be found in the supplemental appendix.
All implementations will be publicly available in our code repository.

\subsection{Configurable Flexibility Through Modularization}
\label{sec:exp_flexibility}
NavComposer’s modular architecture supports flexible component replacement, accommodating mathematical methods, learning-based approaches, and multimodal large models. 
We evaluate various module implementations through controlled experiments, starting with the basic variant \texttt{vo-llv-llv-llm}. 

\subsubsection{Action Classification Module}
We evaluate three implementations of the action classification module: \texttt{vo}, which employs visual odometry to mathematically measure relative motion; and \texttt{dn} and \texttt{rn}, which are neural networks with different backbones.

As shown in Table~\ref{tab:module_action}, the results reveal notable trade-offs among the implementations. 
\texttt{vo} and \texttt{dn} demonstrate comparable performance, with only minor differences across most metrics. 
\texttt{rn} excels in certain aspects of semantic consistency and linguistic diversity but significantly underperforms in contrastive matching.
Further analysis indicates that \texttt{rn} frequently predicts the “move forward” action, contributing to its low ASC score (3.181) and action accuracy (67.3\%). 
This tendency suggests that \texttt{rn} emphasizes entities other than actions, leading to higher SSC and OSC scores.
However, action prediction accuracy remains our primary concern.
Considering both efficiency and accuracy, we select \texttt{vo} as the action classification module due to its well-balanced performance.

\subsubsection{Scene Recognition Module}
We evaluate six scene recognition implementations, as outlined in Table \ref{tab:module_implementations}. 
Among them, \texttt{mae} and \texttt{swg} are unimodal models that process only image inputs, whereas \texttt{llv}, \texttt{blp}, \texttt{qwn}, and \texttt{gpt} are multimodal models capable of processing both images and text.

The results in Table~\ref{tab:module_scene} reveal that smaller unimodal models can perform comparably to certain multimodal models. 
For example, \texttt{mae} achieves similar MAP and CR scores to \texttt{llv}. 
However, large models like \texttt{blp} and \texttt{qwn} outperform unimodal models in matching and consistency metrics, demonstrating their ability to generate richer scene descriptions.
Among all implementations, \texttt{gpt} achieves the best overall performance.
Moreover, we observed that moderate-length scene descriptions tend to yield better evaluation results. 
Overly long descriptions, such as those generated by \texttt{blp}, can degrade instruction quality.
Interestingly, the linguistic diversity is relatively consistent across variants.
This suggests that the indoor scenes in the validation dataset lack substantial scene variety, which constrains vocabulary usage.
Based on these findings, we select \texttt{gpt} as the optimal scene recognition module due to its superior descriptive capabilities, with \texttt{qwn} serving as the best offline alternative.

\begin{table*}[tbp]
\centering
\caption{Performance of different object detection module implementations. Position Rate column indicates the percentage of successfully described object positions. \textbf{Bold} indicates the best performance, while \underline{underlined} indicates the second-best.}
\label{tab:module_object}
\begin{tabular}{Lccccccccccccc}
\toprule
   \multirowcenter{2}{*}{\textrm{Variant}} & \multicolumn{4}{c}{Contrastive Matching} & \multicolumn{4}{c}{Semantic Consistency} & \multicolumn{4}{c}{Linguistic Diversity} & \multirowcenter{2}{*}{\makecell{Position \\ Rate (\%)}} \\ \cmidrule(lr){2-5}\cmidrule(lr){6-9}\cmidrule(l){10-13}
   & HR       & MRR      & TIR      & MAP     & ASC     & SSC     & OSC     & MSC     & MATTR      & NGD     & SBLEU$\downarrow$   & CR$\downarrow$      &    \\ \midrule
vo-llv-llv-llm & 0.731 &  0.615 &  0.489 &  0.607 &  3.882 &  \textbf{2.999} &  4.140 &  \underline{3.674} &  0.580 &  0.973 &  0.893   &  5.379 & 100.0   \\
vo-llv-dtr-llm & 0.621 &  0.497 &  0.380 &  0.492 &  3.205 &  2.610 &  4.042 &  3.285 &  0.576 &  0.985 &  0.889   &  5.183 & 73.6   \\
vo-llv-swg-llm & 0.670 &  0.553 &  0.439 &  0.545 &  3.299 &  2.397 &  3.776 &  3.157 &  0.600 &  \textbf{1.158} &  \underline{0.861}   &  5.170 & 0.0   \\
vo-llv-blp-llm & 0.754 &  0.636 &  0.497 &  0.627 &  3.445 &  2.717 &  3.813 &  3.325 &  0.597 &  1.029 &  0.878   &  5.334 & 94.9   \\
vo-llv-qwn-llm & \underline{0.761} &  \underline{0.645} &  \underline{0.506} &  \underline{0.635} &  \underline{3.924} &  \underline{2.794} &  \underline{4.241} &  3.653 &  \underline{0.606} &  1.014 &  0.878   &  \underline{5.006} & 99.6   \\
vo-llv-gpt-llm & \textbf{0.793} &  \textbf{0.689} &  \textbf{0.563} &  \textbf{0.680} &  \textbf{3.994} &  2.633 &  \textbf{4.745} &  \textbf{3.791} &  \textbf{0.631} &  \underline{1.146} &  \textbf{0.848}   &  \textbf{4.828} & 99.7   \\ \bottomrule
\end{tabular}
\end{table*}

\begin{table*}[tbp]
\centering
\caption{Performance of different instruction synthesis module implementations. The last column presents the average word count in an instruction. \textbf{Bold} indicates the best performance, while \underline{underlined} indicates the second-best.}
\label{tab:module_synthesis}
\begin{tabular}{Lccccccccccccc}
\toprule
   \multirowcenter{2}{*}{\textrm{Variant}} & \multicolumn{4}{c}{Contrastive Matching} & \multicolumn{4}{c}{Semantic Consistency} & \multicolumn{4}{c}{Linguistic Diversity} & \multirowcenter{2}{*}{Length} \\ \cmidrule(lr){2-5}\cmidrule(lr){6-9}\cmidrule(l){10-13}
   & HR       & MRR      & TIR      & MAP     & ASC     & SSC     & OSC     & MSC     & MATTR      & NGD     & SBLEU$\downarrow$   & CR$\downarrow$      &   \\ \midrule
vo-llv-llv-llm & \underline{0.731} &  \textbf{0.615} &  \underline{0.489} &  \textbf{0.607} &  3.882 &  2.999 &  \underline{4.140} &  3.674 &  0.580 &  0.973 &  \underline{0.893}   &  5.379 & 80.1   \\
vo-llv-llv-gmm & 0.591 &  0.483 &  0.311 &  0.478 &  \textbf{5.925} &  \underline{3.561} &  3.287 &  \underline{4.258} &  0.609 &  \underline{1.241} &  0.904   &  \underline{5.189} & 35.1   \\
vo-llv-llv-qwn & \textbf{0.732} &  \underline{0.612} &  0.488 &  \underline{0.606} &  \underline{4.675} &  \textbf{3.789} &  \textbf{4.398} &  \textbf{4.287} &  \underline{0.612} &  \textbf{1.844} &  \textbf{0.797}   &  \textbf{4.629} & 53.8   \\
vo-llv-llv-gpt & 0.714 &  0.608 &  \textbf{0.505} &  0.604 &  4.313 &  3.013 &  3.862 &  3.729 &  \textbf{0.622} &  0.816 &  0.916   &  5.823 & 58.4   \\ \bottomrule
\end{tabular}
\end{table*}

\subsubsection{Object Detection Module}
We evaluate six object detection modules, including unimodal methods (\texttt{dtr}, \texttt{swg}) and multimodal large models (\texttt{llv}, \texttt{blp}, \texttt{qwn}, \texttt{gpt}), as listed in Table~\ref{tab:module_implementations}.

The results in Table~\ref{tab:module_object} clearly demonstrate that multimodal large models significantly outperform unimodal models trained on single datasets. 
For instance, \texttt{dtr} (trained on COCO, 91 object categories)  struggles to generalize to unseen environments, and \texttt{swg} (trained on ImageNet) covers more categories but remains limited in open-domain detection.
In contrast, multimodal models such as \texttt{qwn} excel due to large-scale pre-training, which enables broader object detection capabilities.

\begin{figure}[tbp]
    \centering
    \includegraphics[width=0.48\textwidth]{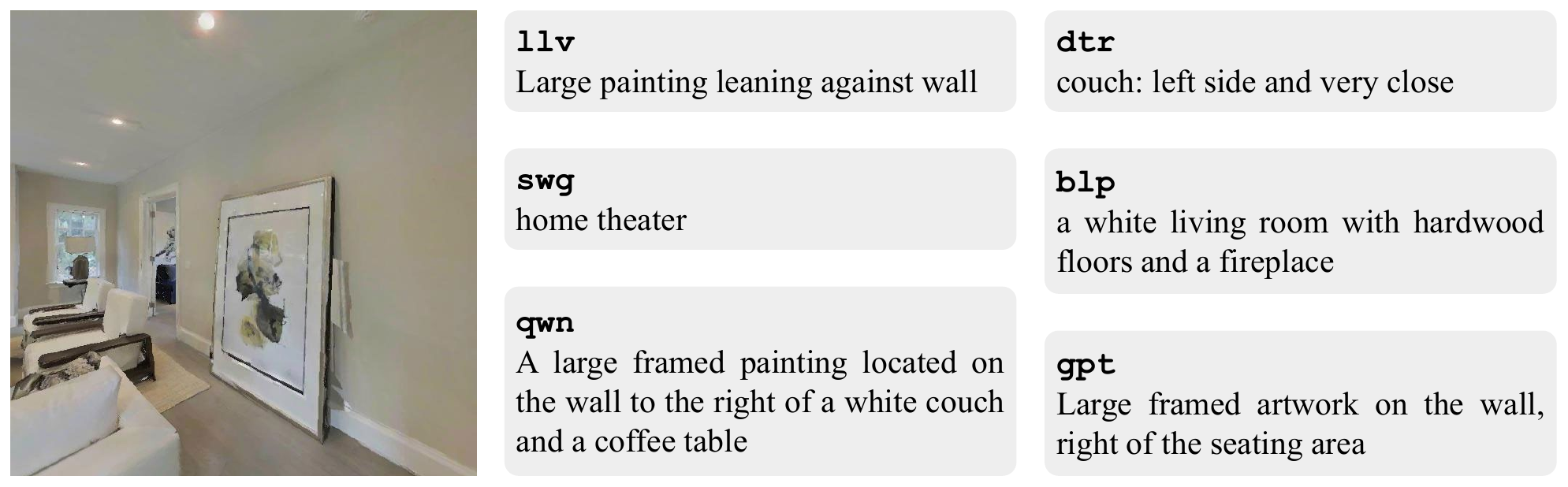}
    \caption{Example outputs of different object detection modules.}
    \label{fig:object_example}
\end{figure}

We also evaluate the success rate of predicting object positions using grammar-based rules (detecting adpositions like ``at," ``on," etc.). 
As expected, models with higher positioning rate generally achieve better overall scores.
For example, \texttt{gpt} attains a 99.7\% success rate in object position prediction, consistent with its strong overall performance. 
Fig. \ref{fig:object_example} visualizes several detection results, illustrating differences in how models describe object positions.
Based on these findings, we select \texttt{gpt} for its high performance and detailed object descriptions, with \texttt{qwn} serving as the best offline alternative.

\subsubsection{Instruction Synthesis Module}
For the instruction synthesis module, we compare four state-of-the-art large language models: \texttt{llm}, \texttt{gmm}, \texttt{qwn}, and \texttt{gpt}. 
The specific versions of these models are listed in Table~\ref{tab:module_implementations}.

Table~\ref{tab:module_synthesis} reveals substantial performance differences among the models for our task.
\texttt{gmm} performs the worst in the matching evaluation, as it generates overly concise instructions focused primarily on action verbs (e.g., ``Turn left, move forward, turn right...").
This lack of descriptive detail results in an abnormally high ASC score but weakens descriptions of other entities.
\texttt{llm} produces the longest instructions but underperforms \texttt{qwn}, indicating that our evaluation metrics are not biased by longer sentence length alone.
Surprisingly, \texttt{gpt} does not achieve the expected performance.
We hypothesize that, being designed for general conversational tasks, \texttt{gpt} may incorporate hidden system prompts and partially overlook the specific characteristics required for navigation instruction synthesis.
Given these findings, we select \texttt{qwn} as the optimal instruction synthesis module due to its strong overall performance and balanced instruction length.

\begin{table*}[tbp]
\caption{Performance comparison of different methods. The upper part lists other methods, while the lower part includes our method and several ablation variants. \textbf{Bold} indicates the best performance.}
\label{tab:method_comparison}
\centering
\begin{tabular}{llcccccccccccc}
\toprule
    & \multirowcenter{2}{*}{\textrm{Method}} & \multicolumn{4}{c}{Contrastive Matching} & \multicolumn{4}{c}{Semantic Consistency} & \multicolumn{4}{c}{Linguistic Diversity} \\ \cmidrule(lr){3-6}\cmidrule(lr){7-10}\cmidrule(l){11-14}
   \# & & HR       & MRR      & TIR      & MAP     & ASC     & SSC     & OSC     & MSC     & MATTR      & NGD     & SBLEU$\downarrow$   & CR$\downarrow$  \\ \midrule
   
\rownumber & Baseline & 0.788 &  0.652 &  0.475 &  0.641 &  2.324 &  3.432 &  2.547 &  2.768 &  0.488 &  0.364 &  0.961 &   8.076 \\

\rownumber & Speaker-Follower \cite{fried2018speaker} & 0.726 &  0.563 &  0.376 &  0.553 &  2.223 &  2.507 &  2.044 &  2.258 &  0.612 &  0.948 &  0.793 &   4.962 \\

\rownumber & EnvDrop \cite{tan2019learning} & 0.769 &  0.608 &  0.421 &  0.599 &  2.352 &  2.766 &  2.181 &  2.433 &  0.567 &  0.745 &  0.847 &   5.835 \\

\rownumber & InternVideo2.5 \cite{wang2025internvideo2} & 0.731 &  0.602 &  0.412 &  0.594 &  2.411 &  3.897 &  2.838 &  3.049 &  0.545 &  0.513 &  0.954 &   7.825 \\

\rownumber & LLaVA-Video \cite{zhang2024llavanext-video} & 0.602 &  0.440 &  0.275 &  0.436 &  2.408 &  3.377 &  2.329 &  2.705 &  0.362 &  0.411 &  0.958 &  10.264  \\

\rownumber & Qwen2.5-VL \cite{Qwen2.5-VL} & 0.799 &  0.685 &  0.522 &  0.675 &  2.532 &  3.467 &  2.617 &  2.872 &  \textbf{0.717} &  0.701 &  0.923 &   6.161 \\ \midrule

\rownumber & NavComposer (ours) &  \textbf{0.802} &  \textbf{0.695} &  0.574 &  \textbf{0.685} &  4.396 &  5.630 &  5.309 &  \textbf{5.112} &  0.670 &  \textbf{1.630} &  \textbf{0.735} &   4.478 \\

\rownumber & -\ w/o action &  0.744 &  0.670 &  \textbf{0.584} &  0.664 &  2.201 &  5.547 &  \textbf{6.508} &  4.752 &  0.666 &  1.488 &  0.816 &   5.327 \\

\rownumber & -\ w/o scene &  0.752 &  0.664 &  0.562 &  0.658 &  4.476 &  1.148 &  5.359 &  3.661 &  0.659 &  1.617 &  0.782 &   4.777 \\

\rownumber & -\ w/o object &  0.712 &  0.615 &  0.524 &  0.611 &  \textbf{5.778} &  \textbf{6.537} &  0.821 &  4.379 &  0.600 &  1.421 &  0.840 &   5.453 \\

\rownumber & -\ offline &  0.788 &  0.669 &  0.529 &  0.659 &  3.542 &  4.362 &  4.946 &  4.283 &  0.674 &  1.594 &  0.762 &   \textbf{4.401} \\
\bottomrule 
\end{tabular}
\end{table*}

\subsection{Comparative Evaluation of Instruction Quality}
\label{sec:exp_quality}
\subsubsection{Method Comparison}
In Table~\ref{tab:method_comparison}, we compare NavComposer (\#7) with several alternative methods using the metrics defined in NavInstrCritic. 
The methods are grouped into three categories for systematic comparison.

The first category consists of an end-to-end baseline (\#1), specifically designed for our evaluation system and introduced in Section~\ref{sec:exp_setup}.  
This baseline serves as a foundational reference for assessing more advanced approaches. 
While \#1 benefits from in-domain training, particularly with contrastive matching, it underperforms NavComposer without task-specific training across all metrics.
This demonstrates NavComposer's ability to generate highly accurate and semantically rich instructions.

\begin{figure}[tbp]
\centering
    \subfloat[]{\includegraphics[width = 0.48\linewidth]{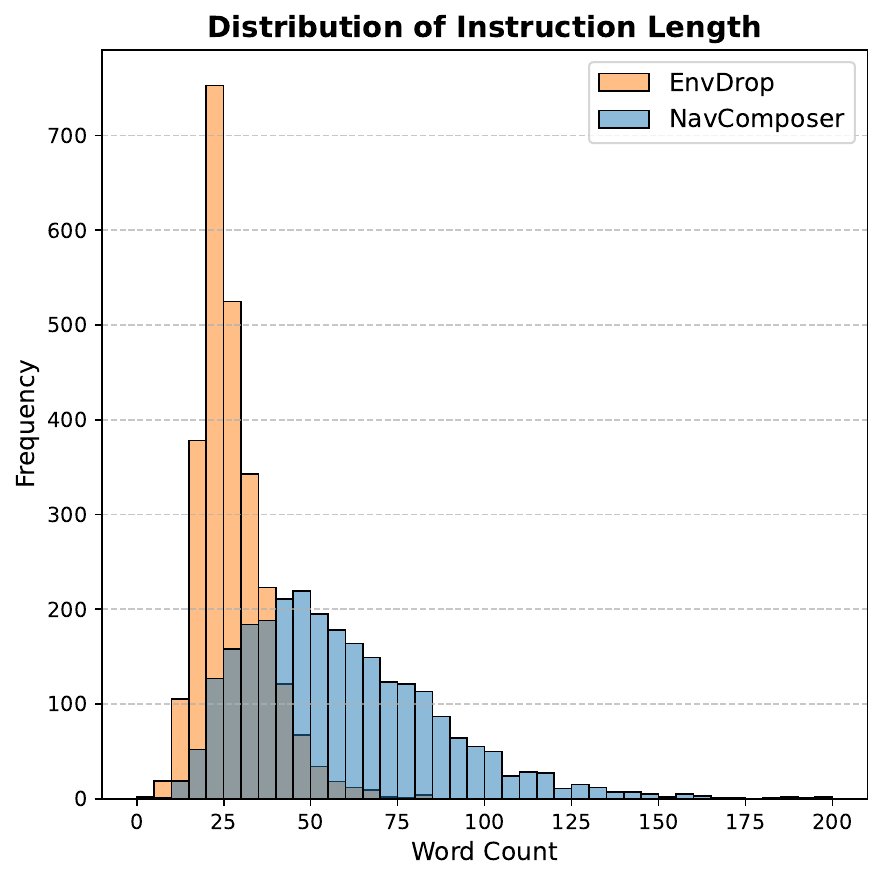}\label{fig:length_distribution}}
    \hfill
    \subfloat[]{\includegraphics[width = 0.50\linewidth]{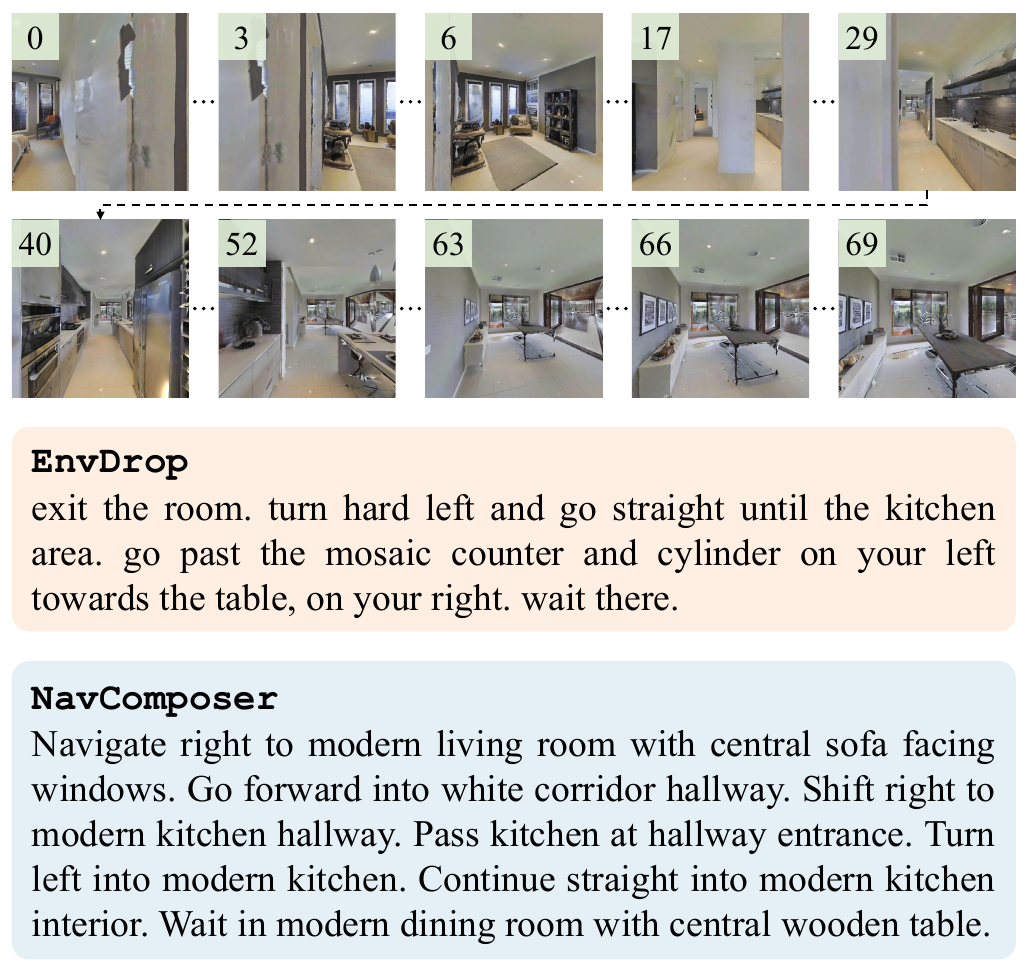}\label{fig:length_example}}
\caption{Instruction lengths and examples generated by NavComposer and EnvDrop. (a) The distribution of instruction lengths, measured by word count. (b) A qualitative example: EnvDrop misunderstands the initial turning direction and omits some details, while our method not only correctly describes actions but also provides richer guidance.}
\label{fig:length_comparison}
\end{figure}

The second category includes classical speaker methods from discrete VLN, such as Speaker-Follower (\#2) and EnvDrop (\#3).
Originally developed for the R2R dataset, their results can be adapted to VLN-CE by mapping shared trajectory IDs. 
Despite being transferable, we find that they fall short in continuous domain evaluation.
Both contrastive matching and semantic consistency evaluations indicate their instruction quality is lower than the baseline.
Their main advantage lies in linguistic diversity, likely due to the richer panoramic vision in discrete settings.
Nevertheless, they remain inferior to NavComposer across all metrics, as NavComposer leverages the capabilities of LLM-based modules.
Additionally, learning-based models (\#1, \#2, \#3) are constrained by training data, producing shorter instructions (27 words on average).
NavComposer can generate more detailed instructions averaging 58 words, as shown in Fig.~\ref{fig:length_comparison}.
These findings highlight the importance of studying instruction generation in continuous navigation domains, rather than relying on adaptations from discrete settings.

The third category includes LLMs with video-processing capabilities, namely InternVideo2.5 (\#4), LLaVA-Video (\#5), and Qwen2.5-VL (\#6). 
Among them, \#5 performs the worst, generating instructions that fail to align with navigation trajectories and exhibit repetitive patterns.
Model \#6 performs better than the baseline, with contrastive matching performance comparable to NavComposer.
However, its expression of semantic entities remains insufficient, achieving only an MSC score of 2.947.
These results highlight the limitations of relying solely on large models for first-person navigation tasks.

In summary, NavComposer outperforms all other methods across nearly all metrics, demonstrating its superiority.
Its modular architecture, explicit extraction of navigation entities, and dedicated focus on continuous navigation domains enable the generation of high-quality, diverse, and informative instructions.
Additional visualizations are provided in Appendix C.

\subsubsection{Ablation Study}
Table~\ref{tab:method_comparison} also presents an ablation study evaluating the impact of different semantic entities in NavComposer.
After removing specific entities (\#8, \#9, \#10), we observe notable performance drops.

Eliminating object entities (\#10) causes the largest decline in contrastive matching and linguistic diversity, underscoring the importance of objects as key navigation landmarks.
Interestingly, ablating scene entities (\#9) significantly reduces SSC but increases ASC and OSC scores 
This reveals an inherent balance in the consistency evaluation: when one type of entity is absent, the model compensates by emphasizing others, leading to score redistribution.
Despite this, all ablated variants yield lower MSC scores than the full model, confirming the value of each entity extraction.

We also evaluate an offline variant \texttt{vo-qwn-qwn-qwn} (\#11), which runs locally and avoids the use of internet APIs.
Its performance is slightly suboptimal to the best-performing version, emphasizing the importance of a powerful LLM backend for optimal results. 
However, the offline variant offers advantages in cost, stability, and time efficiency due to its local deployment and possible acceleration. 
Given these advantages, we use this variant for large-scale experiments in Section~\ref{sec:exp_adaptability}.

\subsubsection{NavInstrCritic Metrics}
The preceding experiments highlight several insights for effectively using the NavInstrCritic metrics.

For contrastive matching, HR and MRR rank the most relevant instruction for a given trajectory.
TIR enhances comprehensiveness by considering all matching positives, while MAP offers the most holistic measure of overall matching performance.
For semantic consistency, ASC, SSC, and OSC evaluate the alignment between entities in the trajectory and instruction. 
These scores often balance each other: removing one entity type may lower its corresponding score while improving others.
MSC, the average of the three, serves as a robust overall indicator, though individual scores can be used for targeted analysis.
For linguistic diversity, MATTR measures token-level variation; NGD and SBLEU assess n-gram diversity at corpus and sentence level, respectively; CR provides an information-theoretic compression efficiency.

Overall, MAP, MSC, and CR are recommended for quickly identifying top-performing models. 
However, different metrics highlight distinct performance aspects, as evidenced by varied rankings in Tables~\ref{tab:module_action} and~\ref{tab:method_comparison}. 
When scores are close, careful metric selection is essential for nuanced analysis.

\begin{table*}[tbp]
    \centering
    \caption{Characteristics and evaluation results for different trajectory sources. ``Instr. Quantity" indicates the number of instructions generated for each source.}
    \label{tab:diverse_generation}
    \begin{tabular}{lcccccccc}
    \toprule
    Source & Device & Domain & Resolution & Steps & Instr. Quantity & MAP & MSC & CR \\
    \midrule
    HM3D \cite{ramakrishnan2021hm3d} & virtual camera & indoor, reconstructed & $224\times 224$ & 60.35 & 30,000 & \textbf{0.62} & 4.10 & 4.38 \\
    SceneNet \cite{mccormac2017scenenet} & virtual camera & indoor, synthesized & $224\times 224$ & 55.51 & 30,000 & 0.28 & 4.23 & 4.21 \\
    GRUtopia \cite{wang2024grutopia} & bipedal robot & indoor, reconstructed & $256\times 256$ & 65.47 & 2,121 & 0.53 & 4.00 & 4.46 \\
    SUN3D \cite{xiao2013sun3d} & handheld camera & indoor, real & $224\times 224$ & 45.66 & 8,484 & 0.60 & 4.41 & 4.11 \\
    ScanNet \cite{dai2017scannet} & handheld camera & indoor, real & $1296\times968$ & 50.64 & 1,959 & 0.55 & 4.61 & 4.24 \\
    TUM \cite{sturm2012benchmark} & handheld camera & indoor, real & $640\times 480$ & 38.66 & 507 & 0.19 & 4.94 & 4.19 \\
    12-Scenes \cite{valentin2016learning} & handheld camera & indoor, real & $1296\times 968$ & 30.49 & 375 & 0.37 & 5.02 & \textbf{4.04} \\
    DIODE \cite{diode_dataset} & tripod-mounted camera & mixied, real & $1024\times 768$ & 22.67 & 3,471 & 0.30 & 5.27 & 4.23 \\
    KITTI \cite{Geiger2013IJRR} & vehicle-mounted camera & outdoor, real & $1240\times 375$ & 54.02 & 492 & 0.23 & \textbf{5.49} & 4.29 \\
    NavTJ & wheeled robot & indoor, real & $640\times 480$ & 146.49 & 114 & 0.26 & 3.75 & 4.19 \\
    \bottomrule
    \end{tabular}
\end{table*}
\begin{figure}[tbp]
\centering
\includegraphics[width=0.48\textwidth]{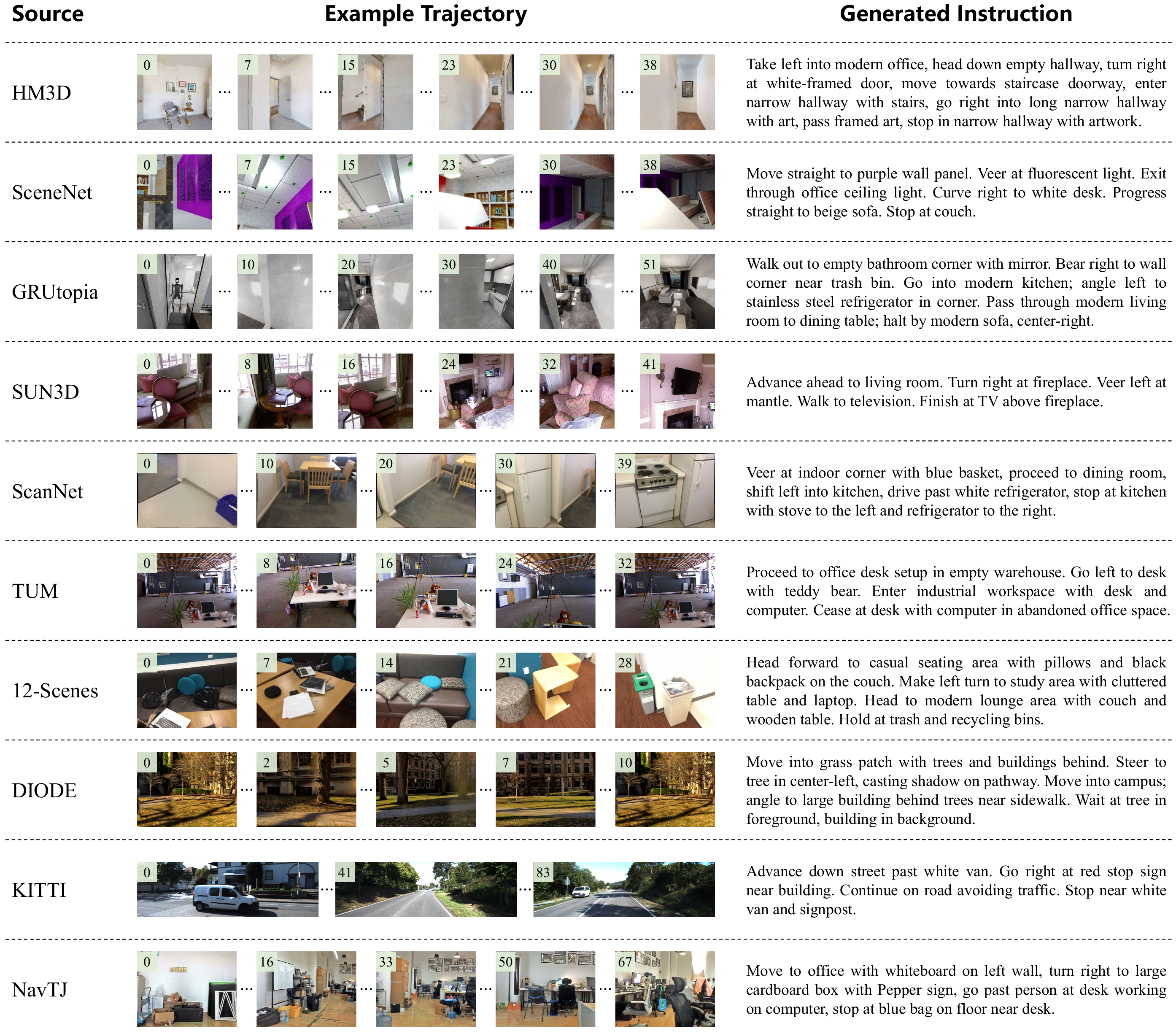}
\caption{Visualization of navigation instructions generated by NavComposer for diverse trajectories.}
\label{fig:various_trajectory}
\end{figure}

\subsection{Universal Adaptability Across Diverse Sources}
\label{sec:exp_adaptability}
Previous experiments primarily used VLN-CE dataset. 
Here, we evaluate NavComposer’s adaptability across 10 datasets spanning diverse devices, domains, and scales (Table~\ref{tab:diverse_generation}).
We only applied minor preprocessing adjustments: long trajectories were segmented, high-frame-rate videos were downsampled for action classification, and each source maintains a maximum of 10,000 trajectories.

As shown in Fig.~\ref{fig:various_trajectory}, NavComposer performs robustly across all sources.
For indoor simulator scenes like HM3D and SceneNet, it captures both room-specific attributes (``narrow hallway," ``purple wall panel") and key landmarks (``framed art," ``beige sofa").
Even in challenging GRUtopia scenarios, with a bipedal robot-mounted shaking camera, it accurately describes transitions such as from a mirrored bathroom to a living room with a table and sofa.
In real-world settings, NavComposer adapts well to handheld camera datasets like SUN3D, ScanNet, TUM, and 12-Scenes.
For instance, in ScanNet, it successfully describes a left turn from a corner toward a stove and refrigerator.
NavComposer also performs well in outdoor environments.
With respect to DIODE, it identifies grass patches, trees, and buildings, inferring the scene to be a campus.
In KITTI, it can process super-wide images and recognizes actions like ``go right" and ``avoid traffic."
Finally, in our lab’s NavTJ dataset, NavComposer generates detailed instructions referencing objects like a whiteboard, Pepper box, computer, and blue bag.
These results affirm NavComposer’s adaptability across varied devices, domains, and resolutions.

To quantitatively assess performance, we calculated MAP, MSC, and CR scores (Table~\ref{tab:diverse_generation}).
The results show that MSC and CR remain effective across all datasets.
Notably, some scores even exceed those on VLN-CE, e.g., KITTI’s MSC and 12-Scenes’ CR.
While cross-dataset comparisons are not entirely equitable, these findings suggest that diverse data can potentially improve consistency and diversity. 
However, contrastive matching presents a more complex picture.
The current contrastive matcher is trained on the VLN-CE training set.
It performs well on HM3D (similarily reconstructed), but poorly on divergent datasets like SceneNet (synthetic) and KITTI (outdoor).
This likely reflects a significant domain mismatch rather than a deficiency in instruction quality, emphasizing the importance of a more generalized matcher.

In summary, NavComposer demonstrates universal adaptability when applied to 10 types of trajectories with highly diverse characteristics.
This experiment also reveals a potential limitation of the NavInstrCritic system when applied to different domains, warranting further investigation.

\section{Conclusion}
In this paper, we addressed the challenge of data scarcity in language-guided navigation by introducing NavComposer and NavInstrCritic, two novel frameworks for navigation instruction generation and evaluation.
NavComposer adopts a modular design that supports flexible integration of different module implementations, enabling the use of state-of-the-art techniques.
Its explicit use of semantic entities significantly enhances the accuracy and quality of the generated instructions.
Operating in a data-agnostic manner, NavComposer demonstrates adaptability across diverse scenarios, overcoming the domain dependencies of traditional methods.
Complementing this, NavInstrCritic evaluates navigation instructions across three dimensions: contrastive matching, semantic consistency, and linguistic diversity.
This holistic evaluation system aligns with the multifaceted nature of navigation descriptions and eliminates reliance on human-annotated ground truth.
Extensive experiments validate the effectiveness of our method, demonstrating significant improvements over existing approaches.
Decoupling instruction generation and evaluation from specific navigation agents may also encourage further research on egocentric task description.

Despite these advancements, our work has certain limitations that warrant future research.
First, although NavComposer supports controllable instruction styles, achieving precise control remains challenging.
An important direction is to develop finer strategies that enable granularity-aware adaptation and improved output customization.
Second, the current version of NavInstrCritic uses a contrastive matcher trained on VLN-CE, which may inherit biases from its limited training data.
Future work could explore using a more robust video-text model to initialize the contrastive matcher.
Finally, the high computational cost of large-scale instruction generation poses challenges for resource-constrained applications.
We aim to optimize deployment on real robots using a cloud-edge framework and explore techniques such as quantization to improve efficiency.

\bibliographystyle{IEEEtran}
\bibliography{references}


 



\vfill

\end{document}